\documentclass[conference]{IEEEtran}
\IEEEoverridecommandlockouts
\usepackage{cite}
\usepackage{amsmath,amssymb,amsfonts}
\usepackage{algorithmic}
\usepackage{graphicx}
\usepackage{textcomp}
\usepackage{xcolor}
\def\BibTeX{{\rm B\kern-.05em{\sc i\kern-.025em b}\kern-.08em
    T\kern-.1667em\lower.7ex\hbox{E}\kern-.125emX}}
\begin{document}

\title{Intra- and Inter-modal Context Interaction Modeling for Conversational Speech Synthesis}

\author{
\IEEEauthorblockN{Zhenqi Jia}
\IEEEauthorblockA{\textit{Inner Mongolia University} \\
Hohhot, China \\
jiazhenqi7@163.com}
\and

\IEEEauthorblockN{Rui Liu* \thanks{* Corresponding Author.}}
\IEEEauthorblockA{\textit{Inner Mongolia University} \\
Hohhot, China \\
liurui\_imu@163.com}

% \thanks{ xxx xxx xxx xxx xxx xxx xxx xxx xxx xxx xxx xxx xxx xxx xxx  }

}

% \author{\IEEEauthorblockN{1\textsuperscript{st} Given Name Surname}
% \IEEEauthorblockA{\textit{dept. name of organization (of Aff.)} \\
% \textit{name of organization (of Aff.)}\\
% City, Country \\
% email address or ORCID}

% \and
% \IEEEauthorblockN{2\textsuperscript{nd} Given Name Surname}
% \IEEEauthorblockA{\textit{dept. name of organization (of Aff.)} \\
% \textit{name of organization (of Aff.)}\\
% City, Country \\
% email address or ORCID}

% \and
% \IEEEauthorblockN{3\textsuperscript{rd} Given Name Surname}
% \IEEEauthorblockA{\textit{dept. name of organization (of Aff.)} \\
% \textit{name of organization (of Aff.)}\\
% City, Country \\
% email address or ORCID}

% \and
% \IEEEauthorblockN{4\textsuperscript{th} Given Name Surname}
% \IEEEauthorblockA{\textit{dept. name of organization (of Aff.)} \\
% \textit{name of organization (of Aff.)}\\
% City, Country \\
% email address or ORCID}

% \and
% \IEEEauthorblockN{5\textsuperscript{th} Given Name Surname}
% \IEEEauthorblockA{\textit{dept. name of organization (of Aff.)} \\
% \textit{name of organization (of Aff.)}\\
% City, Country \\
% email address or ORCID}

% \and
% \IEEEauthorblockN{6\textsuperscript{th} Given Name Surname}
% \IEEEauthorblockA{\textit{dept. name of organization (of Aff.)} \\
% \textit{name of organization (of Aff.)}\\
% City, Country \\
% email address or ORCID}

\maketitle

\begin{abstract}

% Conversational Speech Synthesis (CSS) aims to effectively use dialogue history to infer and generate speech with appropriate conversational prosody. Dialogue is an interactive process, and multimodal dialogue history (MDH) includes text and audio modalities. Previous CSS work typically models the entire MDH implicitly to obtain conversational prosody. However,  intra-modal and inter-modal interaction modeling can more effectively capture the subtle semantic and prosodic variations within the MDH. However, previous work ignores this point. Therefore, we propose a novel Intra-modal and Inter-modal Context Interaction Modeling CSS model, termed I$^{3}$-CSS, that consists of two main components: 1) Intra-Modal and Inter-Modal Interaction Module: Uses contrastive learning with interaction enhancement mechanisms to model intra-modal and inter-modal interactions within the MDH to infer the semantic and prosodic features of the target utterance. 2) Speech Synthesizer: Integrates the inferred semantic and prosodic features into the target utterance to generate speech with appropriate conversational prosody. Experiments show that I$^{3}$-CSS can synthesize more appropriate conversational prosody compared to baseline models. Code and audio samples are available at \textcolor[rgb]{0.93,0.0,0.47}{https://github.com/xxx/xxxx}.

Conversational Speech Synthesis (CSS) aims to effectively take the multimodal dialogue history (MDH) to generate speech with appropriate conversational prosody for target utterance. The key challenge of CSS is to model the interaction between the MDH and the target utterance. Note that text and speech modalities in MDH have their own unique influences, and they complement each other to produce a comprehensive impact on the target utterance. 
Previous works did not explicitly model such intra-modal and inter-modal interactions.
To address this issue, we propose a new intra-modal and inter-modal context interaction scheme-based CSS system, termed I$^{3}$-CSS. 
Specifically, in the training phase, we combine the MDH with the text and speech modalities in the target utterance to obtain four modal combinations, including Historical Text-Next Text, Historical Speech-Next Speech, Historical Text-Next Speech, and Historical Speech-Next Text. Then, we design two contrastive learning-based intra-modal and two inter-modal interaction modules to deeply learn the intra-modal and inter-modal context interaction. In the inference phase, we take MDH and adopt trained interaction modules to fully infer the speech prosody of the target utterance's text content.
Subjective and objective experiments on the DailyTalk dataset show that I$^{3}$-CSS outperforms the advanced baselines in terms of prosody expressiveness. Code and speech samples are available at \textcolor[rgb]{0.93,0.0,0.47}{https://github.com/AI-S2-Lab/I3CSS}.

\end{abstract}


\section{Introduction}
Conversational Speech Synthesis (CSS) aims to take multimodal dialogue history (MDH) to generate speech with appropriate conversational prosody \cite{guo2021conversational}. As Human-Computer Interaction (HCI) systems become increasingly prevalent in our daily lives, CSS has become a crucial component of intelligent interaction systems \cite{zhou2020design, seaborn2021voice, mctear2022conversational}, playing an important role in areas such as virtual assistants, smart customer service, and smart home devices.

% Traditional CSS methods typically model the entire multimodal dialogue history (MDH), which can be categorized into two groups: \textbf{1) RNN-based Dialogue History Modeling}: Guo et al. \cite{guo2021conversational} introduce a coarse-grained context encoder based on RNN, which extracts semantic information from dialogue history. Hu et al. \cite{hu2022fctalker} and Xue et al. \cite{xue2023m} explore fine-grained and coarse-grained RNN-based context dependencies to extract semantic and prosodic information from multimodal dialogue history. Deng et al. \cite{deng2024concss} first model multimodal dialogue history using an RNN-based context encoder and then apply contrastive learning to constrain the discriminability of context features. \textbf{2) GNN-based Dialogue History Modeling}: Li et al. \cite{li2022enhancing} construct the intra-speaker and inter-speaker relationships in dialogue history as a graph and use a GNN to learn speaking style. Li et al. \cite{li2022inferring} use a multi-scale relational graph convolutional network to infer speaking style from multimodal dialogue history. Liu et al. \cite{liu2024emotion} construct the multi-source knowledge of dialogue history into a heterogeneous graph to infer the target utterance's emotion.

Traditional methods for modeling MDH in CSS can be divided into two categories: 1) RNN-based Dialogue History Modeling \cite{guo2021conversational, lee2023dailytalk,xue2023m,deng2024concss, liu2024emphasis};
% Guo et al. \cite{guo2021conversational} propose an RNN-based coarse-grained context encoder to extract semantic information from independent dialogue history. Hu et al. \cite{hu2022fctalker} and Xue et al. \cite{xue2023m} extend the RNN-based context encoder to multiscale and multimodal, but still only extract historical semantics and prosody from independent MDH. Deng et al. \cite{deng2024concss} first independently model MDH and then use contrastive learning to constrain the contextual features of each modality, improving their discriminability.
% \textbf{
2) GNN-based Dialogue History Modeling \cite{li2022enhancing,li2022inferring,liu2024emotion}.
% }: Li et al. \cite{li2022enhancing} construct independent MDH as a graph representing intra-speaker and inter-speaker relationships and use a GNN to learn the speaking style of the target utterance. Li et al. \cite{li2022inferring} adopt a multi-scale relational graph convolutional network to infer the speaking style of the target utterance from independent MDH. Liu et al. \cite{liu2024emotion} construct the multi-source knowledge of the target utterance and MDH as a heterogeneous graph, implicitly modeling the relationship between the target utterance and MDH to infer conversational emotion.
Previous works often model the MDH independently, neglecting its interactions with the target utterance, as \cite{xue2023m, deng2024concss, li2022enhancing, li2022inferring}. We also note that some studies \cite{liu2024emotion} combine the target utterance with the MDH to model their intrinsically complex relationship. However, the key of CSS is to model the interaction between the MDH and the target utterance \cite{lin-etal-2024-advancing}. The above works fail to explicitly explore the deep interactions between the MDH and the target utterance, especially intra-modal and inter-modal interactions between them. 

Text and speech modalities in MDH have their own unique influences, which complement each other to produce a comprehensive impact on the target utterance \cite{mocanu2023multimodal, li2023imf, mao-etal-2021-dialoguetrm-exploring, meng2024masked}.  The prosodic style of the dialogue history conveys rich information beyond the text itself and can even alter the semantics of the target utterance, and different conversational prosody elicits responses with varying prosody \cite{castro-etal-2019-towards, lin-etal-2024-advancing}. Similarly, the semantics of the dialogue history include not only the target utterance's semantics but also the prosodic style. For example, in a conversation, someone might say, ``I lost my pen." If they speak in an agitated tone, the response could be defensive, such as ``I didn’t take your pen," whereas a sad tone might elicit comfort, with responses like ``Let’s look for it together." This demonstrates that the prosodic style of the dialogue history not only influences the prosody of the target utterance but also affects its semantics. Similarly, if the tone remains the same but the semantics differ, such as ``Did you take my pen?" versus ``Did you see my pen?"—despite the same tone, the former might provoke a defensive response like ``Why would I take your pen?" while the latter may invite a more helpful response, such as ``I haven't seen it, but I can help you find it." This shows that differences in conversational semantics also influence the semantics and prosody of the target utterance. Therefore, how to model the intra-modal and inter-modal interaction explicitly for CSS to improve speech expressiveness is the focus of this work.

% Previous work implicitly models the entire MDH to infer the semantics and prosody of the target utterance. However, in the MDH, modeling the intra-modal interactions between historical and subsequent utterances within the same modality, as well as the inter-modal interactions between historical and subsequent utterances across different modalities, can effectively capture semantic and prosodic changes on a per-utterance basis \cite{mao-etal-2021-dialoguetrm-exploring, meng2024masked}. Intra-modal interactions help accurately infer the semantic and prosodic features of the subsequent utterance, while inter-modal interactions leverage the complementarity of modality features, preserving each modality's characteristics while enhancing them with features from the other modality \cite{mocanu2023multimodal, li2023imf}. This is crucial for accurately inferring the semantics and prosody of the target utterance.

\begin{figure*}[ht]
    \vspace{-3.5mm}
    \centering    \includegraphics[width=1\textwidth]{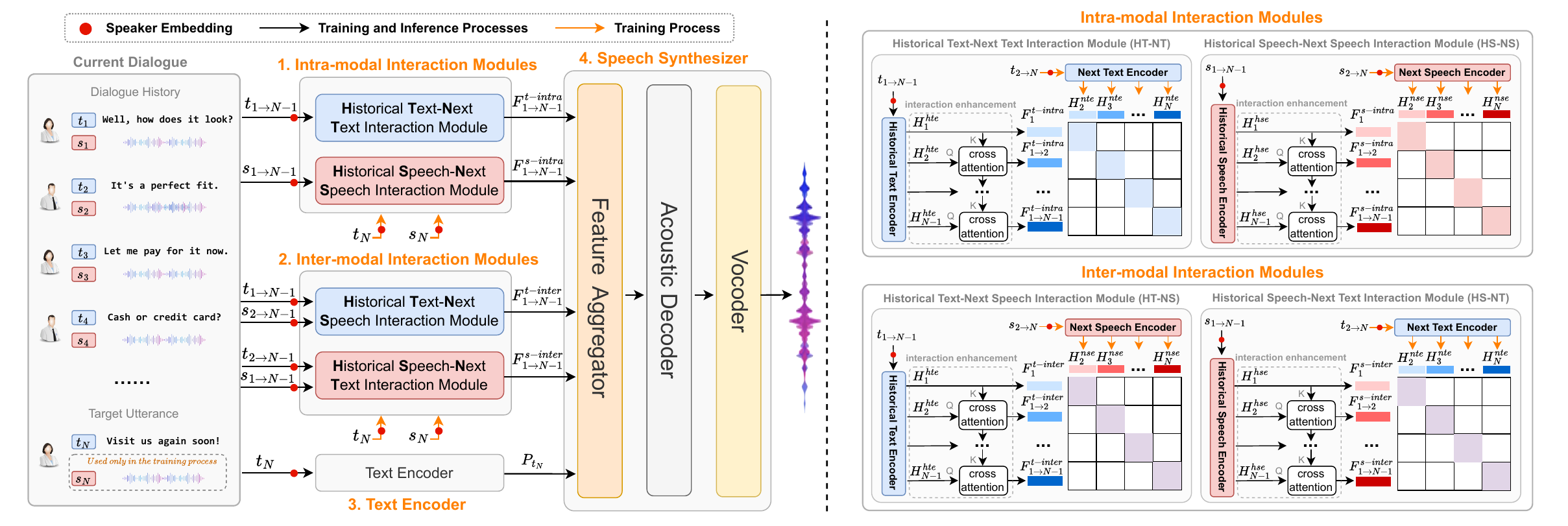}

    % \vspace{-2mm}
    
    \caption{The overview of I$^{3}$-CSS consists of Intra-modal Interaction Modules, Inter-modal Interaction Modules, Text Encoder, and Speech Synthesizer.}

    \vspace{-5mm}
    
    \label{fig1}
\end{figure*}

To address this issue, we propose an \textbf{I}ntra-modal and \textbf{I}nter-modal Context \textbf{I}nteraction scheme-based CSS model, termed I$^{3}$-CSS. Specifically, during the training phase, we design four interaction modules to model the intra-modal and inter-modal interactions between the MDH and the target utterance's text and speech modalities, including Historical Text-Next Text, Historical Speech-Next Speech, Historical Text-Next Speech, and Historical Speech-Next Text. These four modules adopt a contrastive learning-based interaction enhancement mechanism to deeply learn the intra-modal and inter-modal context interactions within the MDH, aligning the semantics and prosody of the target utterance. In the inference phase, we take the MDH and the trained intra-modal and inter-modal interaction modules to fully infer the appropriate conversational prosody for the target utterance. In summary, the main contributions of this paper are as follows:

\begin{itemize}
\item We propose a novel I$^{3}$-CSS model for MDH modeling. To the best of our knowledge, this is the first CSS system to explore intra-modal and inter-modal interactions between MDH and the target utterance.

\item We introduce a contrastive learning-based interaction enhancement method to better learn semantics and prosody variations when modeling intra-modal and inter-modal interactions between MDH and the target utterance.

\item Subjective and objective experimental results show that I$^{3}$-CSS synthesizes speech with more appropriate conversational prosody and naturalness compared to the baseline models.
\end{itemize}

% \begin{table}[htbp]
% \caption{Table Type Styles}
% \begin{center}
% \begin{tabular}{|c|c|c|c|}
% \hline
% \textbf{Table}&\multicolumn{3}{|c|}{\textbf{Table Column Head}} \\
% \cline{2-4} 
% \textbf{Head} & \textbf{\textit{Table column subhead}}& \textbf{\textit{Subhead}}& \textbf{\textit{Subhead}} \\
% \hline
% copy& More table copy$^{\mathrm{a}}$& &  \\
% \hline
% \multicolumn{4}{l}{$^{\mathrm{a}}$Sample of a Table footnote.}
% \end{tabular}
% \label{tab1}
% \end{center}
% \end{table}

% \section{Task Definition}
% The multimodal dialogue can be defined as a sequence of utterances $U_ = \{[t_1, s_1], [t_2, s_2], ..., [t_{N-1}, s_{N-1}], [t_N, s_N]\}$, where $\{t_1, t_2, ..., t_{N-1}\}$ represents the text of the dialogue history, and $t_N$ represents the text of the target utterance. $\{s_1, s_2, ..., s_{N-1}\}$ represents the speech of the dialogue history, and $s_N$ represents the speech to be synthesized.  CSS based on Intra-modal and Inter-modal Context Interaction Modeling needs to consider the following points: 1) How to effectively infer the semantics and prosody of the target utterance through intra-modal interaction modeling within the multimodal dialogue history. 2) How to enhance the target utterance's conversational prosody by utilizing modalities' complementarity in the multimodal dialogue history through inter-modal interaction modeling.

\section{METHODOLOGY}

\subsection{Model Overview}

As shown in Fig. \ref{fig1}, the proposed I$^{3}$-CSS consists of four components: 1) Intra-modal Interaction Modules, 2) Inter-modal Interaction Modules, 3) Text Encoder, 4) Speech Synthesizer. The Intra-modal Interaction Modules aim to infer interactive semantics from the text history and interactive prosody from the audio history. The Inter-modal Interaction Modules aim to infer complementary prosody from the text history and complementary semantics from the audio history. 
The Text Encoder extracts linguistic encodings for the target sentence. Then, the Speech Synthesizer adds these interactive features into the linguistic encodings to generate speech with appropriate conversational prosody.

\subsection{Intra-modal Interaction Modules}
As shown on the left side of Fig. \ref{fig1}, the Intra-modal Interaction Modules consist of the \textbf{H}istorical \textbf{T}ext-\textbf{N}ext \textbf{T}ext Interaction Module (\textbf{HT-NT}) and the \textbf{H}istorical \textbf{S}peech-\textbf{N}ext \textbf{S}peech Interaction Module (\textbf{HS-NS}), which are designed to learn the semantics and prosody of intra-modal interactions between the MDH and the target utterance.

\textbf{HT-NT} aims to learn the semantic interactions between the historical text and the next text, enabling the inference of the target utterance's semantics from the text dialogue history. Specifically, in the Historical Text Encoder, we input $t_{1 \rightarrow N-1}$ into pre-trained Sentence-BERT \cite{reimers-gurevych-2019-sentence} to extract the semantic features of each sentence,  then add the speaker embeddings to these semantic features to retain speaker identity,  and then input these semantic features into a Bi-GRU to obtain the historical semantic features $H^{hte}_{1 \rightarrow N-1}$. We apply interaction enhancement to $H^{hte}_{1 \rightarrow N-1}$. Starting from $H^{hte}_{2}$, we use a cross-attention mechanism to interactively fuse the historical semantic features until obtaining the final intra-modal historical semantic interaction features $F^{t-intra}_{1 \rightarrow N-1}$. The Next Text Encoder uses the same structure as the Historical Text Encoder but takes $t_{2 \rightarrow N}$ as input to obtain single-sentence semantic features $H^{nte}_{2 \rightarrow N}$. To enable the inference of the next text's semantics from the historical text, we design a contrastive learning ($\mathcal{L}^{cl}_{htnt}$) that aligns the historical semantics with the next text's semantics. Specifically, the cosine similarities between \{$F^{t-intra}_{1}$, $F^{t-intra}_{1 \rightarrow 2}, ...$, $F^{t-intra}_{1 \rightarrow N-1}$\} and \{$H^{nte}_{2}$, $H^{nte}_{3}$, ..., $H^{nte}_{N}$\} are used to construct the prediction matrix $M_p$. The ground truth matrix $M_{gt}$ is composed of 1 and -1, where the main diagonal elements are 1, and all other elements are -1. The loss function ($\mathcal{L}^{cl}_{htnt}$) constrains the module by minimizing the mean square error (MSE) between $M_{gt}$ and $M_{p}$, enabling the module to accurately infer the target utterance's semantics from the historical text’s intra-modal semantic interaction features ($F^{t-intra}_{1 \rightarrow N-1}$). 

\textbf{HS-NS} aims to learn the prosody interactions between the historical speech and the next speech, enabling the inference of the target utterance's prosody from the speech dialogue history. The structure of HS-NS is nearly identical to that of HT-NT, except that in the Historical Speech Encoder and the Next Speech Encoder, we use Wav2Vec 2.0 \cite{baevski2020wav2vec} fine-tuned on IEMOCAP \cite{busso2008iemocap} to extract sentence-level prosodic features. Specifically, we input $s_{1 \rightarrow N-1}$ into the Historical Speech Encoder to obtain the historical prosodic features $H^{hse}_{1 \rightarrow N-1}$, then apply interaction enhancement to obtain the intra-modal historical prosody interaction features \{$F^{s-intra}_{1}$, $F^{s-intra}_{1 \rightarrow 2}, ...$, $F^{s-intra}_{1 \rightarrow N-1}$\}. We input $s_{2 \rightarrow N}$ into the Next Speech Encoder to obtain single-sentence prosodic features $H^{nse}_{2 \rightarrow N}$. Finally, the module applies contrastive learning ($\mathcal{L}^{cl}_{hsns}$) enabling it to infer the target utterance' prosody from the historical speech's intra-modal prosody interaction features ($F^{s-intra}_{1 \rightarrow N-1}$).

% During the inference phase, the trained HT-NT and HS-NS are used to extract the two intra-modal interaction features, $F^{t-intra}_{1 \rightarrow N-1}$ and $F^{s-intra}_{1 \rightarrow N-1}$.

\subsection{Inter-modal Interaction Modules}
The Inter-modal Interaction Modules consists of the \textbf{H}istorical \textbf{T}ext-\textbf{N}ext \textbf{S}peech Interaction Module (\textbf{HT-NS}) and the \textbf{H}istorical \textbf{S}peech-\textbf{N}ext \textbf{T}ext Interaction Module (\textbf{HS-NT}), which are used to learn the semantics and prosody inter-modal interactions between the MDH and the target utterance.

\textbf{HT-NS} aims to learn the interaction between the semantics of historical text and the prosody of the next speech, enabling the inference of the target utterance's prosody from the semantics of the text dialogue history. Specifically, $t_{1 \rightarrow N-1}$ is input into the Historical Text Encoder and processed with interaction enhancement to extract the inter-modal historical semantic interaction features \{$F^{t-inter}_{1}$, $F^{t-inter}_{1 \rightarrow 2}$, ..., $F^{t-inter}_{1 \rightarrow N-1}$\}. $s_{2 \rightarrow N}$ is input into the Next Speech Encoder to extract the single-sentence prosodic features $H^{nse}_{2 \rightarrow N}$. Finally, through contrastive learning ($\mathcal{L}^{cl}_{htns}$), the module is constrained to infer the target utterance' prosody from the historical text' inter-modal semantic interaction features ($F^{t-inter}_{1 \rightarrow N-1}$). 

\textbf{HS-NT} aims to learn the interactions between the prosody of historical speech and the semantics of the next text, enabling the inference of the target utterance's semantics from the prosody of the speech dialogue history. Specifically, $s_{1 \rightarrow N-1}$ is input into the Historical Speech Encoder and processed with interaction enhancement to extract the inter-modal historical prosody interaction features \{$F^{s-inter}_{1}$, $F^{s-inter}_{1 \rightarrow 2}$, ..., $F^{s-inter}_{1 \rightarrow N-1}$\}. $t_{2 \rightarrow N}$ is then input into the Next Text Encoder to extract the single-sentence semantic features ($H^{nte}_{2 \rightarrow N}$). Finally, through contrastive learning ($\mathcal{L}^{cl}_{hsnt}$), the module is constrained to infer the target utterance' semantics from the historical speech’s inter-modal prosody interaction features ($F^{s-inter}_{1 \rightarrow N-1}$).

% During the inference phase, the trained HT-NS and HS-NT are used to extract the two inter-modal interaction features, $F^{t-inter}_{1 \rightarrow N-1}$ and $F^{s-inter}_{1 \rightarrow N-1}$.

\subsection{Text Encoder}
The Text Encoder aims to extract phoneme-level linguistic encodings for the target utterance. Specifically, we input $t_N$ into G2P \cite{bisani2008joint} to obtain a phoneme sequence and then use the TTS Encoder from FastSpeech2 \cite{ren2021fastspeech} to extract the linguistic encodings $P_{t_N}$.

\subsection{Speech Synthesizer}
The Speech Synthesizer consists of the Feature Aggregator, the  Acoustic Decoder, and the  Vocoder. The Feature Aggregator adds the four interaction features $F^{t-intra}_{1 \rightarrow N-1}$, $F^{s-intra}_{1 \rightarrow N-1}$, $F^{t-inter}_{1 \rightarrow N-1}$, and $F^{s-inter}_{1 \rightarrow N-1}$ into the linguistic  encodings $P_{t_N}$. The Acoustic Decoder includes a length regulator and a variance adapter to predict duration, energy, and pitch, followed by a mel-decoder to predict mel-spectrogram features. Finally, a pre-trained HiFi-GAN \cite{kong2020hifi} vocoder is used to generate conversational speech.

\subsection{Training and Inference}

During the training phase, we input MDH and the text and audio modalities of the target utterance into the intra-modal and inter-modal interaction modules, and learn the intra-modal and inter-modal interactions between MDH and the target utterance through a contrastive learning-based interaction enhancement mechanism. In the inference phase, we only input MDH into the trained interaction modules to fully infer the speech prosody of the target utterance's text content. 

\begin{table*}[t!]
\centering

\vspace{-4mm}

\caption{Main Results. Green font marks an improvement over the best baseline. I$^3$-CSS  outperforms all baselines with p-value $<$ 0.001.}
\vspace{-2mm}

\resizebox{1\linewidth}{!}{
\begin{tabular}{lccccc}
\hline
\textbf{Systems}  & \textbf{N-DMOS} ($\uparrow$)  & \textbf{P-DMOS} ($\uparrow$)   & \textbf{MAE-P} ($\downarrow$)   & \textbf{MAE-E} ($\downarrow$)   & \textbf{MAE-D} ($\downarrow$)     \\ \hline

DailyTalk \cite{lee2023dailytalk}     & 3.620 $\pm$ 0.028    & 3.640 $\pm$ 0.029    & 0.530   & 0.467   & 0.204     \\

M$^2$-CTTS \cite{xue2023m}     & 3.683 $\pm$ 0.029    & 3.690 $\pm$ 0.029    & 0.543   & 0.380   & 0.146   \\

CONCSS \cite{deng2024concss}    & 3.719 $\pm$ 0.028    & 3.752 $\pm$ 0.031    & 0.482   & 0.328   & 0.143   \\

Homogeneous Graph-based CSS \cite{li2022inferring}    & 3.727 $\pm$ 0.028    & 3.726 $\pm$ 0.030    & 0.489   & 0.320   & 0.146   \\

ECSS \cite{liu2024emotion}     & 3.743 $\pm$ 0.028    & 3.772 $\pm$ 0.030    & 0.505   & 0.332   & 0.134   \\ \hline

% \textbf{I$^{3}$-CSS (Proposed)}  & \textbf{3.864 $\pm$ 0.029} \textcolor[rgb]{0.0, 0.52, 0.52}{\(  (+0.121) \)}   & \textbf{3.876 $\pm$ 0.026} \textcolor[rgb]{0.0, 0.52, 0.52}{\(  (+0.121) \)}    & \textbf{0.450} \textcolor[rgb]{0.0, 0.52, 0.52}{\(  (+0.121) \)}   & \textbf{0.310} \textcolor[rgb]{0.0, 0.52, 0.52}{\(  (+0.121) \)}  & \textbf{0.129} \textcolor[rgb]{0.0, 0.52, 0.52}{\(  (+0.121) \)}     \\ \hline

% \textbf{I$^{3}$-CSS (Proposed)}  & \textbf{3.864 $\pm$ 0.029} \textcolor[rgb]{0.0, 0.52, 0.52}{\scriptsize \( (+0.121) \)}   & \textbf{3.876 $\pm$ 0.026} \textcolor[rgb]{0.0, 0.52, 0.52}{\scriptsize \( (+0.121) \)}    & \textbf{0.450} \textcolor[rgb]{0.0, 0.52, 0.52}{\scriptsize \( (+0.121) \)}   & \textbf{0.310} \textcolor[rgb]{0.0, 0.52, 0.52}{\scriptsize \( (+0.121) \)}  & \textbf{0.129} \textcolor[rgb]{0.0, 0.52, 0.52}{\scriptsize \( (+0.121) \)}     \\ \hline

\textbf{I$^{3}$-CSS (Proposed)}  & \textbf{3.864 $\pm$ 0.029} \textcolor[rgb]{0.0, 0.52, 0.52}{\tiny \( (+0.121) \)}   & \textbf{3.876 $\pm$ 0.026} \textcolor[rgb]{0.0, 0.52, 0.52}{\tiny \( (+0.104) \)}    & \textbf{0.450} \textcolor[rgb]{0.0, 0.52, 0.52}{\tiny \( (+0.032) \)}   & \textbf{0.310} \textcolor[rgb]{0.0, 0.52, 0.52}{\tiny \( (+0.010) \)}  & \textbf{0.129} \textcolor[rgb]{0.0, 0.52, 0.52}{\tiny \( (+0.005) \)}     \\ \hline

\end{tabular}
}

\label{tab1}
\end{table*}

% ------------

% \begin{table*}[t!]
% \centering

% \caption{Subjective (with 95\% confidence interval) and objective results with different CSS systems.}

% \resizebox{1\linewidth}{!}{
% \begin{tabular}{lccccccccccc}
% \hline
% \textbf{Systems}  &  & \textbf{N-DMOS} ($\uparrow$)  &  & \textbf{P-DMOS} ($\uparrow$)   &  & \textbf{MAE-P} ($\downarrow$)   &  & \textbf{MAE-E} ($\downarrow$)   &  & \textbf{MAE-D} ($\downarrow$)     \\ \hline

% DailyTalk \cite{lee2023dailytalk}     &  & 3.620 $\pm$ 0.028    &  & 3.640 $\pm$ 0.029    &  & 0.530   &  & 0.467   &  & 0.204     \\

% M$^2$-CTTS \cite{xue2023m}     &  & 3.683 $\pm$ 0.029    &  & 3.690 $\pm$ 0.029    &  & 0.543   &  & 0.380   &  & 0.146   \\

% CONCSS \cite{deng2024concss}    &  & 3.719 $\pm$ 0.028    &  & 3.752 $\pm$ 0.031    &  & 0.482   &  & 0.328   &  & 0.143   \\

% Homogeneous Graph-based CSS \cite{li2022inferring}    &  & 3.727 $\pm$ 0.028    &  & 3.726 $\pm$ 0.030    &  & 0.489   &  & 0.320   &  & 0.146   \\

% ECSS \cite{liu2024emotion}     &  & 3.743 $\pm$ 0.028    &  & 3.772 $\pm$ 0.030    &  & 0.505   &  & 0.332   &  & 0.134   \\ \hline

% \textbf{I$^{3}$-CSS (Proposed)}  &  & \textbf{3.864 $\pm$ 0.029}    &  & \textbf{3.876 $\pm$ 0.026}   &  & \textbf{0.450}   &  & \textbf{0.310}  &  & \textbf{0.129}     \\ \hline

% \end{tabular}
% }

% \label{tab1}
% \end{table*}

% ------------

\begin{table*}[t!]
\centering

\vspace{-3mm}

\caption{\label{tab2}Ablation results. ``-'' indicates the removal of a particular module, while ``$\checkmark$'' indicates its retention.}

\vspace{-2mm}

\resizebox{1\linewidth}{!}{

\begin{tabular}{l|cccc|ccccc}
\cline{1-10}
\textbf{Systems} & \multicolumn{4}{c|}{\textbf{Ablation Setup}} & \multicolumn{5}{c}{\textbf{Metrics}} \\
\cline{2-10}
 & HT-NT & HS-NS & HT-NS & HS-NT  & $\text{N-DMOS}$ ($\uparrow$) & $\text{P-DMOS}$ ($\uparrow$) & $\text{MAE-P}$ ($\downarrow$) &  $\text{MAE-E}$ ($\downarrow$) & $\text{MAE-D}$ ($\downarrow$) \\

\cline{1-10}

\textit{Abl.Exp.1} & {-} &{-} & {-}& {-}& 3.598 $\pm$ 0.027 & 3.615 $\pm$ 0.022 & 0.681 & 0.588 & 0.245\\

\textit{Abl.Exp.2} & $\checkmark$ &{-} & {-}& {-}& 3.755 $\pm$ 0.029 & 3.768 $\pm$ 0.034 & 0.475 & 0.318 & 0.132\\

\textit{Abl.Exp.3} &{-} & $\checkmark$ &{-} &{-} & 3.776 $\pm$ 0.025 & 3.788 $\pm$ 0.031 & 0.476 & 0.316 & 0.131 \\

\textit{Abl.Exp.4} & {-}& {-}& $\checkmark$ & {-}& 3.781 $\pm$ 0.031 & 3.751 $\pm$ 0.026 & 0.475 & 0.323 & 0.131  \\

\textit{Abl.Exp.5} &{-} &{-} &{-} & $\checkmark$ & 3.753 $\pm$ 0.029 & 3.757 $\pm$ 0.027 & 0.474 & 0.324 & 0.130 \\

\textit{Abl.Exp.6} & $\checkmark$ & $\checkmark$ & {-}& {-} & 3.791 $\pm$ 0.029 & 3.820 $\pm$ 0.031 & 0.469 & 0.321 & 0.131 \\

\textit{Abl.Exp.7} &{-} &{-} & $\checkmark$ & $\checkmark$ & 3.809 $\pm$ 0.028 & 3.794 $\pm$ 0.027 & 0.464 & 0.318 & 0.132 \\

\textit{Abl.Exp.8} & $\checkmark$ &{-} & $\checkmark$ &{-} & 3.792 $\pm$ 0.028 & 3.817 $\pm$ 0.032 & 0.466 & 0.322 & 0.132 \\
\textit{Abl.Exp.9} &{-} & $\checkmark$ &{-} & $\checkmark$ & 3.797 $\pm$ 0.030 & 3.802 $\pm$ 0.026 & 0.460 & 0.319 & 0.132 \\
Abl.Exp.10: w/o IE & $\checkmark$ & $\checkmark$ & $\checkmark$ & $\checkmark$ & 3.752 $\pm$ 0.029 & 3.747 $\pm$ 0.028 & 0.470 & 0.324 & 0.131 \\

\cline{1-10}

\textbf{I$^{3}$-CSS (Proposed)} & $\checkmark$ & $\checkmark$ & $\checkmark$ & $\checkmark$ & \textbf{3.864 $\pm$ 0.029}   & \textbf{3.876 $\pm$ 0.026}   & \textbf{0.450}   & \textbf{0.310}  & \textbf{0.129} \\
\hline
\end{tabular}

}

\vspace{-4.5mm}

\end{table*}

\section{EXPERIMENTS}

\subsection{Experimental Setup}
We validate I$^{3}$-CSS on the English dialogue dataset DailyTalk \cite{lee2023dailytalk}. DailyTalk contains 2,541 dialogues performed by one male and one female speaker. The dataset includes 23,773 speech segments, totaling 20 hours. We split the data into training, validation, and test sets with a ratio of 8:1:1.

In I$^{3}$-CSS, the intra-modal and inter-modal interaction modules share the weights of the historical text encoder, next text encoder, historical speech encoder, and next speech encoder. The historical text encoder and next text encoder use the same network structure, where the speaker embedding dimension is 512, the input channels of the bidirectional GRU are 512, and the output channels are 512. The bidirectional GRU concatenates the forward and backward hidden states into 1024 dimensions, which is then reduced to 256 dimensions through two linear layers. The historical speech encoder and next speech encoder also use the same network structure, with a speaker embedding dimension of 768, the bidirectional GRU input channels set to 768, and output channels set to 768. The bidirectional GRU concatenates the forward and backward hidden states into 1536 dimensions, which is then reduced to 256 dimensions through two linear layers. We use the Adam optimizer with $\beta_1 = 0.9$ and $\beta_2 = 0.98$. We use the Grapheme-to-Phoneme (G2P) \cite{bisani2008joint} toolkit to convert all text inputs into phoneme sequences for text processing. Montreal Forced Alignment (MFA) \cite{mcauliffe17_interspeech} is used to extract phoneme durations and perform alignment. All speech samples are resampled to 22.05 kHz, and Mel spectrogram features are extracted with a 25ms window length and 10ms hop length. The training of I$^{3}$-CSS is conducted on a single A100 GPU with a batch size of 16 and 400k training steps.

\subsection{Evaluation Metrics}
For the subjective evaluation metrics, we organize a dialogue-level mean opinion score (DMOS) \cite{streijl2016mean} listening test with 20 graduate students whose second language is English and provide all volunteers with professional training on the rules. We ask volunteers to first listen to the historical dialogue speech and then to the synthesized target speech to rate the Naturalness DMOS (N-DMOS) and Prosody DMOS (P-DMOS) on a scale of 1 to 5. Please note that N-DMOS focuses on quality and naturalness, while P-DMOS focuses on appropriate conversational prosody. 

For the objective evaluation metrics, we calculate the mean absolute error (MAE) between the predicted and ground truth acoustic features to assess the prosody performance of the synthesized speech. Specifically, we use MAE-P, MAE-E, and MAE-D to evaluate the accuracy of pitch, energy, and duration predictions.

\subsection{Comparative and Ablation Models}
The comparison models are categorized into two groups: \textbf{1) RNN-based dialogue history modeling}: DailyTalk \cite{lee2023dailytalk} models the independent dialogue history through a coarse-grained context encoder to enhance speech expressiveness. M$^{2}$-CTTS \cite{xue2023m} designs a multimodal, multiscale context encoder to model independent MDH to generate speech with appropriate prosody. CONCSS \cite{deng2024concss} increases the discriminability of independent MDH context through contrastive learning. \textbf{2) GNN-based dialogue history modeling}: Homogeneous Graph-based CSS \cite{li2022inferring} uses a homogeneous graph to model independent MDH, inferring speaking styles of the target utterance. ECSS \cite{liu2024emotion} constructs a heterogeneous graph of the target utterance and MDH's multi-source knowledge to predict emotions and synthesize emotionally expressive speech.

For the ablation models, Abl.Exp.1 represents the removal of all intra-modal and inter-modal interaction modules. Abl.Exp.2 to Abl.Exp.9 represents different combinations of HT-NT, HS-NS, HT-NS, and HS-NT, with detailed combinations shown in Table \ref{tab2}. Abl.Exp.10 represents I$^{3}$-CSS without the interaction enhancement (IE) mechanism.

\subsection{Main Results}

As shown in Table \ref{tab1}, I$^{3}$-CSS achieves the best overall performance compared to baselines. In terms of subjective metrics, I$^{3}$-CSS outperforms all baselines in both N-DMOS (3.864) and P-DMOS (3.876). For the objective metrics, it also achieves the best results in MAE-P (0.450), MAE-E (0.310), and MAE-D (0.129). Statistical analysis indicates that I$^3$-CSS significantly outperforms the baselines with a p-value less than 0.001. Experimental results show that by modeling the intra-modal and inter-modal interactions between the MDH and the target utterance, I$^{3}$-CSS better captures the subtle semantics and prosody variations, enabling the generation of target speech with appropriate conversational prosody.

\subsection{Ablation Results}
As shown in Table \ref{tab2}, removing different modules from I$^{3}$-CSS leads to a decline in most subjective and objective metrics. Compared to Abl.Exp.1, Abl.Exp.2-5 show significant improvements in both subjective and objective metrics, indicating that adding the HT-NT, HS-NS, HT-NS, or HS-NT modules to model the intra-modal and inter-modal interactions in the MDH effectively enhances the prosody and naturalness of the generated speech. Abl.Exp.6-9 combine any two modules from HT-NT, HS-NS, HT-NS, or HS-NT, and achieve further improvements over Abl.Exp.2-5 in most metrics. This demonstrates that jointly modeling the semantics and prosody interactions in the MDH, while leveraging the complementarity between modalities, enhances the prediction of conversational prosody for the target utterance. In Abl.Exp.9, we remove the interaction enhancement mechanism, and compared to I$^{3}$-CSS, most metrics show a notable decline, further proving the importance of interaction enhancement in capturing and integrating intra-modal and inter-modal interaction.

\section{CONCLUSION AND FUTURE WORK}
To improve the capability of CSS systems in generating speech with appropriate conversational prosody, we propose a novel I$^{3}$-CSS, which explicitly captures intra-modal and inter-modal interactions between multimodal dialogue history (MDH) and the target utterance, inferring semantics and prosody of the target utterance from the MDH to express appropriate conversational prosody. Experimental results demonstrate that I$^{3}$-CSS significantly outperforms the advanced CSS systems in terms of prosody expressiveness. To our knowledge, I$^{3}$-CSS is the first model to use intra-modal and inter-modal interaction modeling of MDH in CSS. We hope this research provides a new perspective for modeling MDH in CSS. In the future, we will model the interactions between multi-scale MDH and the target utterance, and further explore the finer-grained intra-modal and inter-modal interaction effects.

% \section{Acknowledgments}
% This work was funded by the Young Scientists Fund (No. 62206136) and the General Program (No. 62476146) of the National Natural Science Foundation of China, the ``Inner Mongolia Science and Technology Achievement Transfer and Transformation Demonstration Zone, University Collaborative Innovation Base, and University Entrepreneurship Training Base'' Construction Project (Supercomputing Power Project) (No.21300-231510).

% \begin{thebibliography}{00}
% \bibitem{b1} G. Eason, B. Noble, and I. N. Sneddon, ``On certain integrals of Lipschitz-Hankel type involving products of Bessel functions,'' Phil. Trans. Roy. Soc. London, vol. A247, pp. 529--551, April 1955.

% \end{thebibliography}

\bibliographystyle{IEEEtran}
\bibliography{refs}

\end{document}